\documentclass[conference]{IEEEtran}
\IEEEoverridecommandlockouts
\usepackage{cite}
\usepackage{amsmath,amssymb,amsfonts}
\usepackage{algorithmic}
\usepackage{graphicx}
\usepackage{textcomp}
\usepackage{xcolor}
\usepackage{url}
\usepackage{tabularx}
\usepackage{booktabs} 
\usepackage{pifont}
\def\BibTeX{{\rm B\kern-.05em{\sc i\kern-.025em b}\kern-.08em
    T\kern-.1667em\lower.7ex\hbox{E}\kern-.125emX}}
\renewcommand\IEEEkeywordsname{Keywords}

\usepackage[pscoord]{eso-pic}
\newcommand{\placetextbox}[3]{
\setbox0=\hbox{#3}
\AddToShipoutPictureFG*{ \put(\LenToUnit{#1\paperwidth},\LenToUnit{#2\paperheight}){\vtop{{\null}\makebox[0pt][c]{#3}}}
}
}
\placetextbox{.23}{0.055}{\small{979-8-3315-1127-2/24/\$31.00~\copyright 2024 IEEE}}

\begin{document}

\title{AI-Driven Relocation Tracking in Dynamic Kitchen Environments}

\author{
	\IEEEauthorblockN{Arash Nasr Esfahani\IEEEauthorrefmark{1}\IEEEauthorrefmark{2}, Hamed Hosseini\IEEEauthorrefmark{1}, Mehdi Tale Masouleh\IEEEauthorrefmark{1}, Ahmad Kalhor\IEEEauthorrefmark{1}, Hedieh Sajedi\IEEEauthorrefmark{2}}

	\IEEEauthorblockA{\IEEEauthorrefmark{1}Human and Robot Interaction Lab, School of Electrical and Computer Engineering, University of Tehran, Tehran, Iran}
	\IEEEauthorblockA{\IEEEauthorrefmark{2}School of Mathematics, Statistics, and Computer Science, University of Tehran, Tehran, Iran}
}

\maketitle

\begin{abstract}
As smart homes become more prevalent in daily life, the ability to understand dynamic environments is essential which is increasingly dependent on AI systems. This study focuses on developing an intelligent algorithm which can navigate a robot through a kitchen, recognizing objects, and tracking their relocation. The kitchen was chosen as the testing ground due to its dynamic nature as objects are frequently moved, rearranged and replaced.
Various techniques, such as SLAM feature-based tracking and deep learning-based object detection (e.g., Faster R-CNN), are commonly used for object tracking. Additionally, methods such as optical flow analysis and 3D reconstruction have also been used to track the relocation of objects. These approaches often face challenges when it comes to problems such as lighting variations and partial occlusions, where parts of the object are hidden in some frames but visible in others. The proposed method in this study leverages the YOLOv5 architecture, initialized with pre-trained weights and subsequently fine-tuned on a custom dataset. A novel method was developed, introducing a frame-scoring algorithm which calculates a score for each object based on its location and features within all frames. This scoring approach helps to identify changes by determining the best-associated frame for each object and comparing the results in each scene, overcoming limitations seen in other methods while maintaining simplicity in design.
The experimental results demonstrate an accuracy of 97.72\%, a precision of 95.83\% and a recall of 96.84\% for this algorithm, which highlights the efficacy of the model in detecting spatial changes.

\end{abstract}

\IEEEpeerreviewmaketitle
\section{Introduction}
The integration of robots into human environments necessitates their ability to perceive their dynamic surroundings. A crucial element of this capability is the ability to detect changes in the environment. This task requires overcoming substantial challenges related to variations in object appearance, occlusions, and positional changes. Moreover, real-world factors, including variable lighting conditions and complex object arrangements, further complicate these challenges. This complexity calls for more sophisticated systems capable of accurately interpreting dynamic environments.
\begin{figure}
	\centering
	\includegraphics[width=0.5\textwidth]{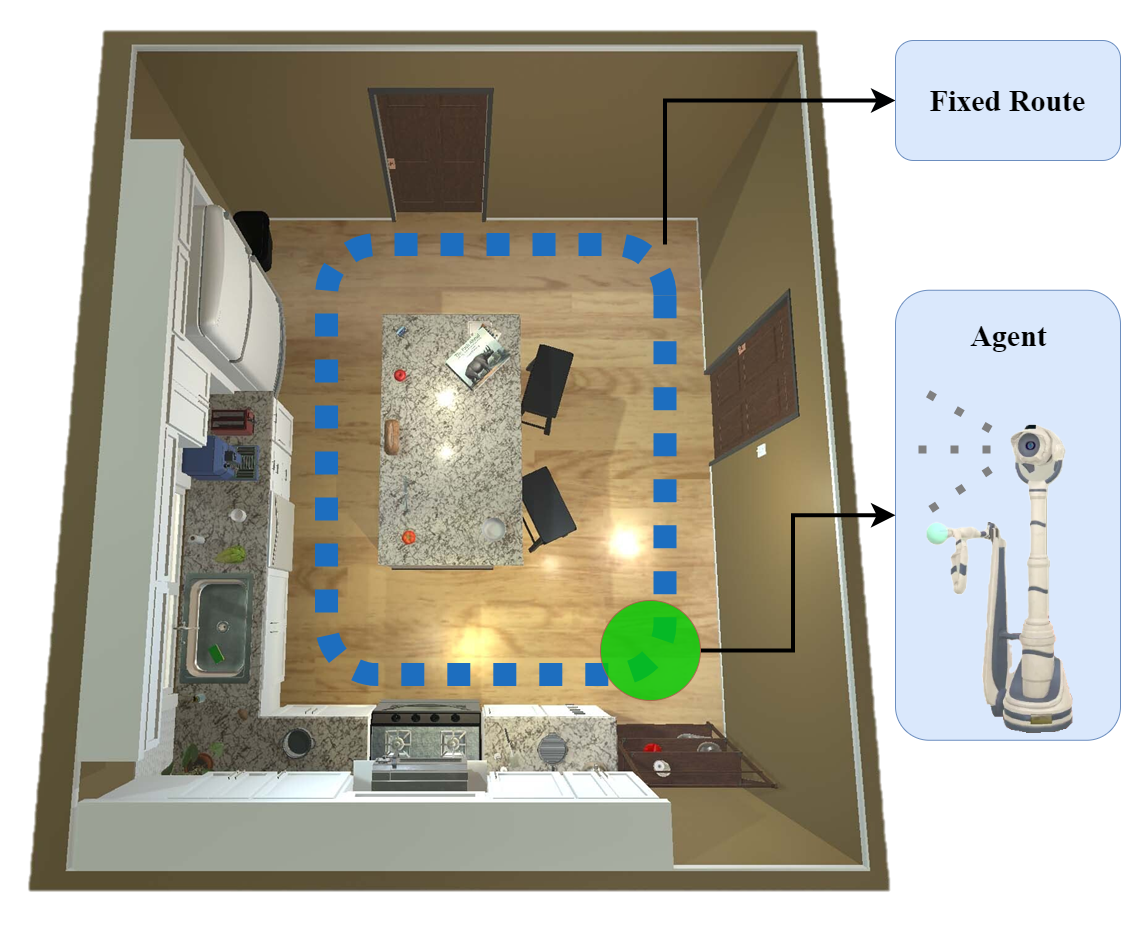}
	\caption{Top view of the kitchen environment, showing the agent's route and the positions of various objects. (Note: This angle was not used during the training or evaluation phases of the project.)}
	\label{fig:angle}
\end{figure}
Despite many advancements in computer vision, traditional object-tracking approaches frequently struggle with partial occlusions \cite{object_relocation_tracking} and changes in object location. These limitations significantly affect the system's effectiveness in automating routine tasks and supporting daily activities. Previous work on object rearrangement, such as \cite{rearrangement} and \cite{simple_approach}, has primarily centered on pick-and-place tasks. Similarly, \cite{room_rearrangement} focuses on accurate placement rather than the detection and tracking of object relocation, which is the primary goal of this research.

Furthermore, few studies have specifically addressed the detection of changes in kitchen environments. When such research has been conducted, it has mainly focused on object detection rather than tracking object relocation, which is a crucial aspect for rearrangement. Notably, \cite{sagues} and \cite{kitchen} primarily emphasize object detection in real-world scenarios while neglecting the importance of recognizing spatial changes. Additionally, in \cite{Ghasemi2024}, spatial relationships in static images are analyzed to identify objects and determine the pick-and-place sequence; however, this work is limited to a fixed view and does not consider dynamic tracking in a broader environment.

In the final stage of rearrangement, the robot should grasp objects, pick them up, and place them at designated locations \cite{hosseini2024multi}\cite{hosseini2020improving}. Some methods utilize robotic grippers to accomplish this task \cite{koosheshi2023agile}. While real-world execution poses challenges in robotic grasping, simulators simplify the pick-and-place process using specialized tools which bypass the complexities of grasping, allowing for a focus on scene understanding, rather than the intricacies of manipulation.

This study introduces an innovative approach to understanding dynamic and cluttered environments, focusing on detecting changes between initial and final object configurations. By combining object detection with a new relocation tracking system, this method identifies the necessary adjustments to transform the initial scene into the final arrangement. The system utilizes the AI2-THOR platform \cite{thor} as the simulator combined with the YOLO object detection model \cite{yolo}, which is trained on diverse kitchen objects.

The main contribution of this paper is creating a novel algorithm called ``best-associated frame selection'' for tracking the relocation of objects. By combining deep learning techniques with a frame-association algorithm, the proposed system demonstrates its effectiveness in accurately identifying and tracking objects in complex scenes.

The research methodology for this study is presented in the following sections. First, Section \ref{sec:Dataset Collection} details the dataset collection approach utilizing a suitable simulation environment. Next, Section \ref{sec:Model Training} describes the selection and training of an object detection model to identify the location and size of objects within each frame of the scene. The simulation setup and performance assessment procedures are outlined in Section \ref{sec:Experimental Setup and Performance Assessment} which describes the navigation approach in the scene in order to track object relocation. A novel algorithm is then introduced for relocation tracking in Section \ref{sec:Best-Associated Frame Algorithm}. Subsequently, Section \ref{sec:Experiments and Results} presents an analysis of the results obtained from both the object detection and relocation tracking algorithms. Finally, the paper concludes in Section \ref{sec:Conclusion} with a summary of the key findings and a discussion of potential future research directions.

\section{Relocation Tracking Approach}
\label{sec:Proposed Approach}
Choosing the suitable simulator is crucial for accurately testing and validating object detection systems, as it directly impacts the realism and variability of the test scenarios. Despite considering ManiSkill2 \cite{mani} and Habitat \cite{habitat}, because of its superior scene complexity, object variety, and interactive features, AI2-THOR was chosen as the simulator for this study. Using AI2-THOR, a dataset was collected by using its built-in capabilities. A YOLOv5 model was then trained on this dataset to detect objects in the simulated environment. After training, the model was tested in a simulated kitchen environment using a fixed route. The agent captured frames before and after scene manipulation and then used the YOLOv5 model for object detection in these captured scenes. This process involves assigning a score to each object and determining the best-associated frame for tracking relocation.

\subsection{Dataset Collection for Dynamic Kitchen Scenes}
\label{sec:Dataset Collection}
The dataset for this study was gathered using the AI2-THOR simulator, which provides a diverse collection of images from simulated kitchen environments. This section details the data collection method, including the use of AI2-THOR's built-in annotation tools and the utilization of data augmentation techniques, along with a thorough description of the key characteristics and properties of the resulting dataset.

\subsubsection{Data Collection Method}
The data collection process involved using AI2-THOR's built-in tool which provides automatic annotation. These annotations included detailed bounding box information for each object instance in a scene. This information was then adjusted to fit YOLOv5's format. Specifically, the original coordinates were transformed into a format that includes the center position, width, and height of the bounding boxes, and these values were normalized to work seamlessly with YOLOv5.

\begin{figure}
	\centering
	\includegraphics[width=0.4\textwidth]{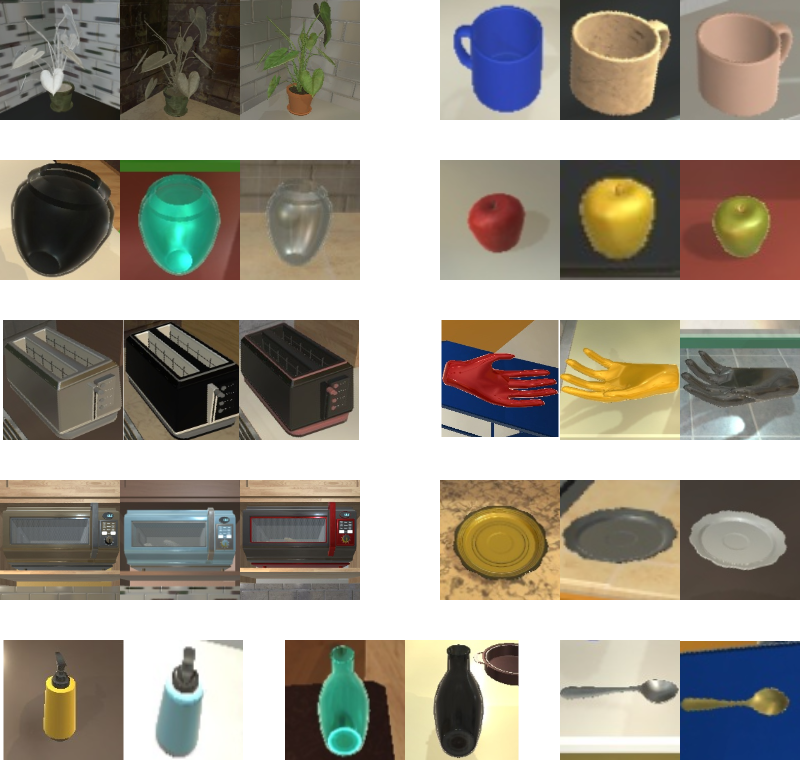}
    \caption{Variability of object properties in the dataset collected using the AI2-THOR simulator.}
	\label{four}
\end{figure}

Data collection took place within simulated kitchen environments, designed to replicate real-world household settings. The process involved systematically positioning the AI agent in various locations within the kitchen, capturing images from multiple angles and distances, and interacting with objects in the environment. In order to further increase the diversity of the dataset, built-in tools were utilized to randomize materials and place objects randomly in the scene which resulted in varied object textures and colors while maintaining their structural properties as seen in Fig. \ref{four}. Through this structured approach, it was ensured that the model was trained on diverse variety of objects and reducing potential biases such as background configurations, and thus enhancing the generalizability of the resulting system.

\subsubsection{Dataset Properties}

The data collection process yielded a dataset of over 9,000 images, each paired with corresponding bounding box annotations for all the objects in the image. The resulting dataset covers a broad spectrum of common kitchen objects, including utensils, food items, and furniture. Examples of objects include refrigerators, microwaves, knives, forks, pots, apples, and chairs. The inclusion of diverse objects across different categories enables the development of a highly generalizable object detection model.

Recent works such as \cite{act_thor,asc_anything,tidee,robot_kitchen_assistance}, highlight the significance of distinguishing between different states of objects (e.g., closed fridge vs. opened fridge) in robotic perception, as these distinctions are crucial for understanding the current state of the environment. Based on these observations, various object states were incorporated into the dataset, with each state treated as a distinct class. This approach resulted in a total of 69 distinct object classes in the dataset with 29 of them being different states of another class. For example, a bowl and a filled bowl would be considered separate classes in this dataset. Each class was represented by at least 225 samples, providing sufficient data for accurate model training.

In the next step, the dataset was preprocessed and augmented. A small portion of the dataset (5\%) was converted to grayscale to simulate diverse lighting conditions, further enhancing the model's robustness. Auto-orientation feature was employed to ensure proper image alignment, and all images were resized to $640\times640$ pixels to match YOLOv5's input requirements. Finally, the dataset was divided into training, validation, and test sets, with 72\% of the images allocated to the training set, 19\% to the validation set, and 9\% to the test set. These preprocessing techniques, combined with the diversity of objects and scene configurations, resulted in a comprehensive training dataset for object detection models.

\subsection{Object Detection}
\label{sec:Model Training}
Following dataset collection, the model training phase can be started.
After analyzing various architectures of YOLO \cite{architecture}, the YOLOv5s model was selected, a compact variant of the YOLOv5 family, for the object detection task. This choice was based on its optimal balance between computational efficiency and detection accuracy. 
The preprocessed dataset was used to train the YOLOv5s model. It was trained for 300 epochs, a duration empirically determined to be sufficient for convergence while reducing the risk of overfitting. A batch size of 16 was employed, adhering to the conventional specifications for YOLOv5 models. The learning rate was initialized at 0.01 and modulated using cosine annealing scheduling, facilitating adaptive learning throughout the training process. In order to enhance the model's robustness and generalization capabilities, several data augmentation techniques were implemented. These included mosaic augmentation, which combines multiple images into a single training instance, as well as random horizontal flipping and rotation. Moreover, color space transformations such as adjustments to brightness, contrast, and hue were also applied to simulate various lighting conditions and improve the model's performance across diverse environments.

\subsection{Simulation Setup and Performance Assessment}
\label{sec:Experimental Setup and Performance Assessment}
After training the model, it was tested in a dynamic kitchen environment. The primary goal was to evaluate the system's ability to detect changes in object positions using the newly developed best-associated frame selection algorithm.
To systematically capture the scene and identify changes, a fixed route was established for the agent within the AI2-THOR simulator. This route could be navigated by executing a series of forward, backward, left, and right movements for the agent. Additionally, the agent could perform head rotations in four directions: left, right, up, and down. Fig. \ref{fig:angle} provides an approximate representation of the agent's route, illustrating the comprehensive coverage of the kitchen environment.

The agent's route is designed to thoroughly inspect the kitchen space. Through iterative experimentation, the route was finalized with strategic stops at corners and key areas. It utilizes zigzag patterns in the central area to capture multiple angles of each object, while also focusing attention on high object density areas, such as countertops. This comprehensive approach ensures a detailed examination of the entire kitchen environment. To accurately track and detect changes in a dynamic environment, a methodical approach is employed where an agent captures and compares images from the same route before and after any changes occur.
\begin{figure*}
	\centering
	\includegraphics[width=1\textwidth]{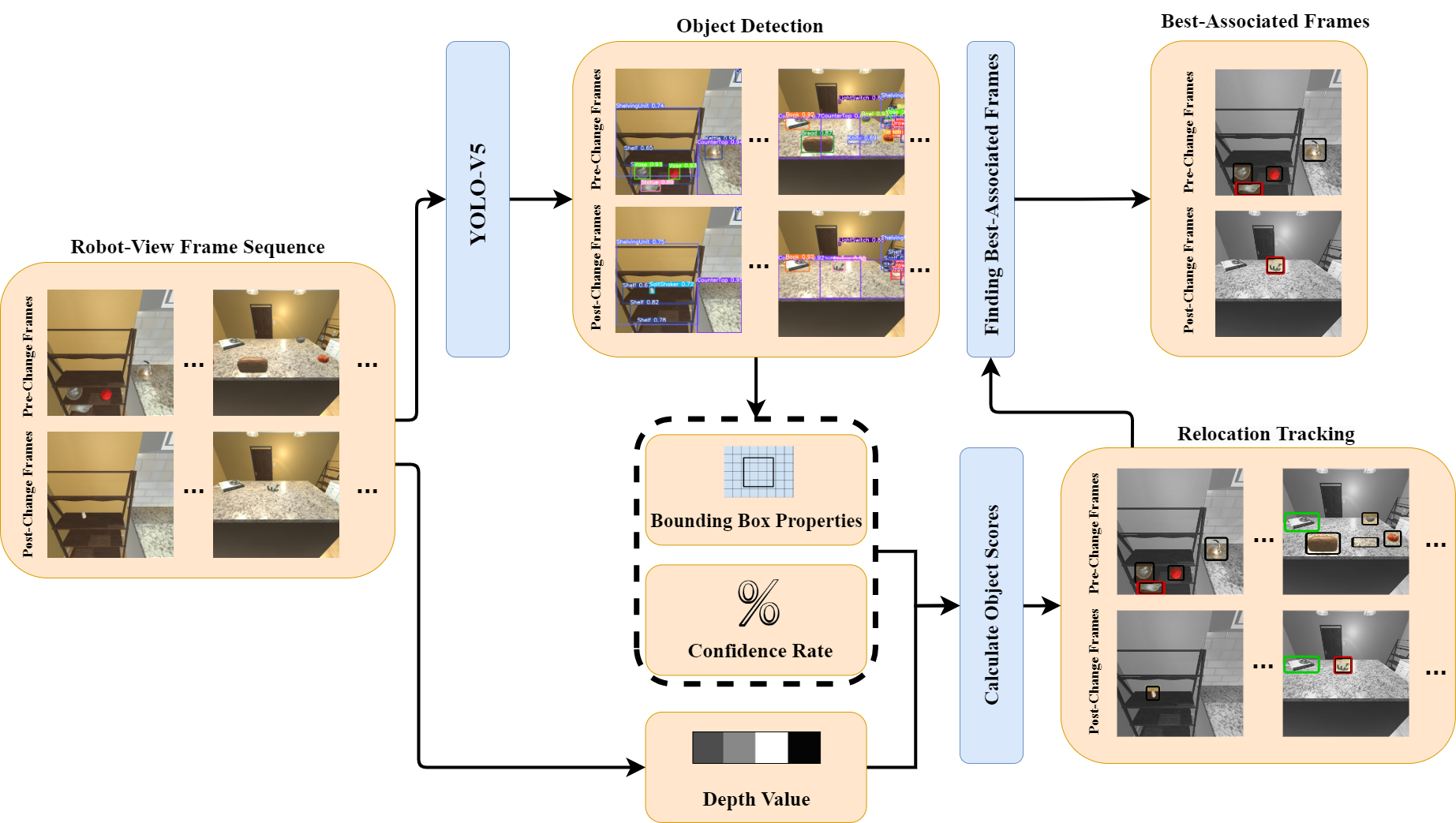}
	\caption{Schematic overview of the proposed methodology. The system computes a score for each object within a frame to identify the best-associated frame. In the relocation tracking phase, green squares indicate no change in object position, black squares indicate a change in position, and red squares indicate a change in object position, highlighting the current frame as the best-associated frame for the relocated object.}
	\label{fig:flow}
\end{figure*}
Pre-change and post-change scenes are essential concepts for relocation tracking in this method. In the pre-change scene, an agent follows a predefined path, capturing images to establish a baseline of the environment. After changes occur, the agent retraces the same route in the post-change scene. This allows for a precise comparison between frames from both scenes, enabling the identification of differences, such as new or missing objects. Maintaining a consistent route is crucial for accurate detection, as it minimizes the effects of lighting or other environmental variations, ensuring reliable analysis of changes in the scene.

In the pre-change scene, the agent navigates this predefined route, capturing images at regular intervals. These images serve as the baseline for subsequent change detection. Following scene modifications, the agent retraces the identical fixed route in the post-change scene. This consistent movement pattern is critical for change detection, as it allows for direct frame-to-frame comparison. The consistency in the agent's route ensures the detection of positional changes from the same perspectives, accurate identification of new or removed objects, and minimal impact from lighting or environmental variations.

As the agent navigates the scene, frames are stored at each movement along the predetermined route. In the AI2-THOR environment, the system captures a frame whenever the agent changes orientation or position. Each frame is annotated using the YOLOv5 model to extract information about objects. The implementation of this fixed route strategy, coupled with the trained YOLOv5 model, forms the foundation of the relocation tracking system. By comparing the detected objects and their spatial relationships between the pre-change and post-change scenes, changes in the kitchen environment can be accurately identified. This approach enables robust performance in dynamic settings, addressing the challenges posed by frequent object movements and rearrangements typical in kitchen scenarios.

\subsection{Best-Associated Frame Selection Algorithm}
\label{sec:Best-Associated Frame Algorithm}
In order to determine the optimal frame for accessing each object, an object scoring algorithm was developed. This algorithm assesses the quality of each frame for each object and identifies the optimal frame for subsequent analysis. The best-associated frame selection algorithm is based on a scoring formula that evaluates an object's visibility in each frame. This algorithm was developed through rigorous experimentation and is defined as follows:

\begin{small}
\begin{equation*}
    \text{Visibility Score} = 2(1 - D) + (10 WH) + (1 - C) + F
\end{equation*}
\end{small}
where:
\begin{itemize}
    \item $D$ is the depth of the object from the agent.
    \item $W$ and $H$ are the width and height of the object's bounding box, respectively.
    \item $C$ is the distance of the bounding box from the frame's center.
    \item $F$ is the confidence rate of the object detection.
    \item All the above values are normalized between zero and one.
\end{itemize}
 
The scoring formula incorporates several key factors to evaluate object visibility. Depth prioritizes objects closer to the agent. Bounding box size puts a greater emphasis on the greater bounding boxes, with the term scaled by a coefficient of 10 to balance with other factors. The centrality factor prioritizes objects located closer to the center of the frame. Finally, confidence gives preference to higher confidence detections extracted from the YOLO model. These factors collectively contribute to a comprehensive assessment of an object's visibility and relevance in each frame.

Following the scoring of objects in the frames, the best-associated frames of each object across both pre-change and post-change scenes are compared. A significant difference is flagged if the distance between the pre-change and post-change frames exceeds a threshold of 9 frames. This threshold was empirically determined through iterative testing, creating a balance between sensitivity to object relocation and robustness against minor variations. The final output of the proposed method is a comprehensive list of objects that have considerably shifted positions, along with their corresponding best-associated frames in both the pre-change and post-change scenes as seen in Fig. \ref{fig:flow}.

\begin{figure*}
	\centering
	\includegraphics[width=1\textwidth]{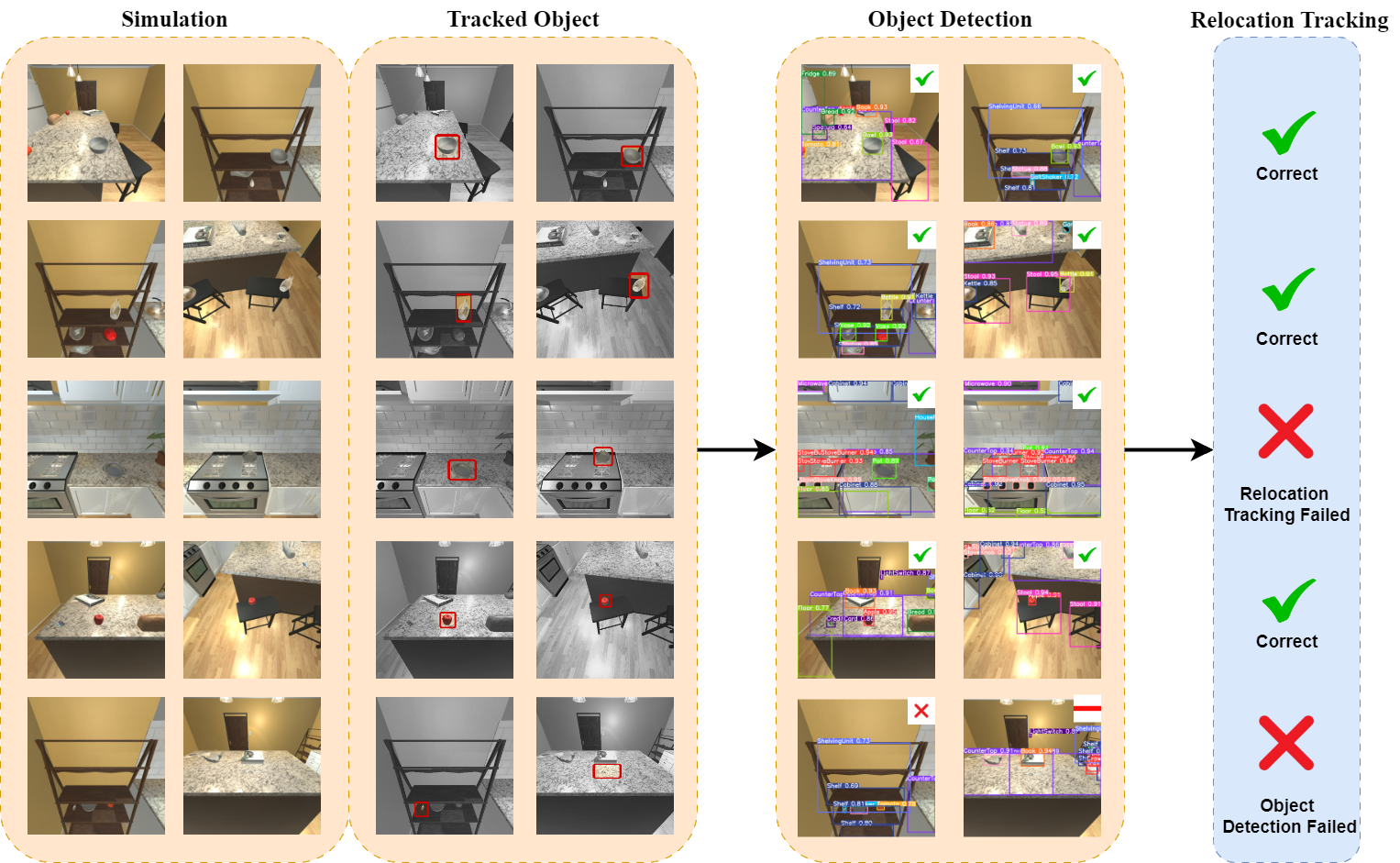}
	\caption{Best-associated frame selection algorithm results of objects in 5 different randomized scenes. The object detection and relocation tracking results are presented separately. Check marks (\ding{51}), hyphens (-), and cross marks (\ding{55}) represent successful detections, missed detections, and unsuccessful detections, respectively.}
	\label{fig:results}
\end{figure*}

\section{Results and Discussion}
\label{sec:Experiments and Results}
In order to assess the efficiency of the proposed algorithm, it is crucial to analyze the results related to object detection. However, the most critical aspect lies in evaluating the algorithm’s effectiveness in tracking object relocation, which will be addressed in the final analysis.

\subsection{Object Detection Results}
Two versions of the model were trained, the baseline model and a refined model that incorporated adjustments to the dataset. The results in Table \ref{tab:comp} compare the performance of the baseline and refined models. The refined model, which yielded the optimal results, incorporated a strategic modification to the dataset by excluding distant objects from each frame. This change led to a noticeable improvement in recall, increasing from 72.5\% to 75.5\%, and a higher mAP50, rising from 78.8\% to 81.6\%. The improvement can be attributed to the fact that YOLOv5 treats the background as a separate class, and by focusing on closer, more distinct objects, the refined model was able to learn more discriminative features. However, this adjustment resulted in a slight reduction in precision, with the refined model achieving 84.8\% compared to the baseline’s 89.6\%. This trade-off suggests that by prioritizing closer objects, the refined model enhanced its ability to distinguish between features, leading to improved overall detection performance despite the slight decrease in precision.

\begin{table}[tp]
    \centering
    \caption{Comparison of Detection Metrics: Baseline and Refined.}
    \begin{tabular}{|l|c|c|}
        \hline
          & \textbf{Baseline Model} & \textbf{Refined Model} \\
        \hline
        \textbf{Precision} & 89.6\%  & 84.8\% \\
        \hline
        \textbf{Recall}  & 72.5\% & 75.5\% \\
        \hline
        \textbf{mAP50}  & 78.8\% & 81.6\%\\
        \hline
    \end{tabular}
    \vspace{0.2cm}
    \label{tab:comp}
\end{table}

\subsection{Relocation Tracking Results}

In order to evaluate the model’s capability in detecting and tracking the relocation of objects, an experiment using 9 randomly generated scenes and a default scene was conducted in the AI2-THOR simulation environment. Each scene contained between 60 and 80 objects, and in total, 614 objects across all scenes were examined to determine whether they had been relocated. The scenes were unbiased and created using AI2-THOR’s randomization tools. The intelligent agent followed a fixed route through the environment, comparing all 9 scenes against the default. The results demonstrate the effectiveness of the best-associated frame selection algorithm in detecting spatial changes of objects in each scene. The confusion matrix for the experiment is presented in Table \ref{tab:confusion_matrix}.

\begin{table}[tp]
    \centering
    \caption{Confusion Matrix for Object Relocation Tracking.}
    \begin{tabular}{|c|c|c|}
        \hline
        & \textbf{Predicted} & \textbf{Predicted} \\
        & \textbf{Relocation} & \textbf{No Relocation} \\
        \hline
        \textbf{Actual Relocation} & \multicolumn{1}{c|}{184} & \multicolumn{1}{c|}{6} \\
        \hline
        \textbf{Actual No Relocation} & \multicolumn{1}{c|}{8} & \multicolumn{1}{c|}{416} \\
        \hline
    \end{tabular}
    \vspace{0.2cm}
    \label{tab:confusion_matrix}
\end{table}

Several instances of these potential results are depicted in Fig. \ref{fig:results}. The performance metrics for the best-associated frame selection algorithm, based on the provided confusion matrix, indicate a precision of 95.8\%, a recall of 96.8\% and an accuracy of 97.7\%. The superior performance of the relocation tracking algorithm compared to the object detection algorithm can be attributed to two key factors. Firstly, the object tracking algorithm only considers objects with a confidence score above 80\%, ensuring that only highly probable detections are used for tracking. Secondly, the system requires the detection of an object in only one frame within a sequence of frames to identify a change in its position. This relaxed requirement allows the algorithm to effectively track objects even if they are temporarily occluded or experience brief periods of low visibility. However, if the relocation is too small, it may go undetected during the relocation tracking process, as it can be observed from Fig. \ref{fig:results}.

\section{Conclusion}
\label{sec:Conclusion}
This study successfully developed and evaluated a system for object relocation tracking within dynamic kitchen environments. By integrating advanced computer vision techniques with the AI2-THOR simulation platform and creating a comprehensive dataset of over 9,000 images across 69 distinct object classes, the proposed system demonstrated high accuracy in detecting and tracking relocated objects. Two iterations of the YOLOv5 model were trained, with the refined model achieving improved results by strategically modifying the training data to focus on closer objects, increasing the mAP50 from 78.8\% to 81.6\%. The best-associated frame selection algorithm significantly outperformed the base object detection model, achieving a precision of 95.8\% and a recall of 96.8\% compared to the  object detection model's 84.8\% precision and 75.5\% recall. This substantial improvement can be attributed to the multi-frame approach, which only considered objects with high confidence scores and required detection in just one frame within a sequence to identify positional changes. While the system has shown excellent performance in kitchen environments, its current application is limited to this single type of room. Future work should aim to extend the system's capabilities to track objects throughout an entire house. This expansion would involve adapting the algorithm to handle a wider range of environments, such as living rooms and bedrooms. These advancements lay the groundwork for robots capable of performing practical tasks and interacting with objects in real-world homes.

\bibliographystyle{IEEEtran}

\end{document}